\documentclass{article}

\usepackage{arxiv}
\usepackage{algorithm}
\usepackage{algorithmic}
\usepackage{float}

\usepackage[utf8]{inputenc} 
\usepackage[T1]{fontenc}    
\usepackage{hyperref}       
\usepackage{url}            
\usepackage{booktabs}       
\usepackage{amsfonts}       
\usepackage{nicefrac}       
\usepackage{microtype}      
\usepackage{lipsum}		
\usepackage{graphicx}
\usepackage{doi}

\title{Efficient Argument Classification with Compact Language Models and ChatGPT-4 Refinements}

\author{ {\hspace{1mm}Marcin Pietron} \\
	Institute of Electronics\\
	AGH University of Krakow\\
	Kraków, Poland \\
	\texttt{pietron@agh.edu.pl} \\
	\And
	{\hspace{1mm}Rafał Olszowski} \\
	Center for Collective Intelligence\\
        MIT Massachusetts Institute of Technology\\
	Cambridge, MA, United States\\
        \And
	{\hspace{1mm}Jakub Gomułka} \\
	Faculty of Humanities\\
	AGH University of Krakow\\
	Kraków, Poland \\
}

\renewcommand{\shorttitle}{\textit{arXiv} Template}

\hypersetup{
pdftitle={A template for the arxiv style},
pdfsubject={q-bio.NC, q-bio.QM},
pdfauthor={David S.~Hippocampus, Elias D.~Striatum},
pdfkeywords={First keyword, Second keyword, More},
}

\begin{document}
\maketitle

\begin{abstract}
Argument mining (AM) is defined as the task of automatically identifying and extracting argumentative components (e.g. premises, claims, etc.) and detecting the existing relations among them (i.e., support, attack, no relations).
Deep learning models enable us to analyze arguments more efficiently than traditional methods and extract their semantics. This paper presents comparative studies between a few deep learning-based models in argument mining. The work concentrates on argument classification. The research was done on a wide spectrum of datasets (Args.me, UKP, US2016). 
The main novelty of this paper is the ensemble model which is based on BERT architecture and ChatGPT-4 as fine tuning model. The presented results show that BERT+ChatGPT-4 outperforms the rest of the models including other Transformer-based and LSTM-based models. 
The observed improvement is, in most cases, greater than 10
The presented analysis can provide crucial insights into how the models for argument classification should be further improved. Additionally, it can help develop a prompt-based algorithm to eliminate argument classification errors. 

\end{abstract}
\section{Introduction}
Argument mining (AM) is a multidisciplinary research field encompassing diverse areas such as logic and philosophy, language, rhetoric and law, psychology, and computer science. The theory of argumentation and the use of logical reasoning to justify claims and conclusions is an extensively studied field, but the application of data science methods to automate these processes is a relatively recent development. In nearly every field, the ability to automatically extract arguments and their relationships from the input source is of significant importance. Over the last decade, AM has become one of the core studies within artificial intelligence \cite{ref_article1,ref_article2} due to its ability to conjugate representational needs with user-related cognitive models and computational models for automated reasoning \cite{ref_article3}. As a subfield of Natural Language Processing (NLP) and computational linguistics, AM focuses on automatically identifying, extracting, and analyzing argumentative structures within natural language texts, which includes recognizing core components of arguments, such as claims and evidence \cite{ref_article4}. Early AM research predominantly focused on edited texts \cite{ref_article5,ref_article6,ref_article7}. However, more recent investigations have broadened their scope to encompass user-generated content found on various platforms, including debate portals, as well as social media sites like Facebook and Twitter \cite{ref_article8}. This expansion demonstrates the growing applicability and relevance of AM in understanding and analyzing diverse forms of discourse across different online platforms.

In AM, the whole task can be decomposed into three main sub-tasks depending on their argumentative complexity. First, the identification of argument components consists in distinguishing argumentative propositions from non-argumentative propositions. This allows segmentation of the input text into arguments, making it possible to carry out the subsequent sub-tasks. Second, the identification of clausal properties is the part of AM that focuses on finding premises or conclusions among the argumentative propositions. Third, the last sub-task is the identification of relational properties. Two different argumentative propositions are considered at a time, and the main objective is to identify which type of relation links both propositions. This task can be solved by a classification model.

The Transformer-based model architecture and its subsequent pre-training approaches have been a turning point in the NLP research area (e.g. BERT, Google T5). Conversely, the emergence of Chat-GPT, a generative pre-trained transformer language model developed by OpenAI, has presented new opportunities for argument mining. Introduced in November 2022, Chat-GPT is based on the GPT architecture and refined via comprehensive training on large-scale datasets. This model can comprehend context, produce coherent and contextually appropriate responses, and participate in meaningful dialogues with users.


This research paper aims to explore the cutting-edge capabilities of the Transformer-based models in the field of Argument mining. Given the rising prominence of Large Language Models (LLMs) such as GPT-4, our research introduces the concept of the 'compact language model' (CLM) as a distinct category. Contrary to LLMs like GPT-4, Microsoft Copilot, or Google's PaLM, CLMs have fewer parameters, which makes them faster to run and easier to deploy in resource-constrained environments. Furthermore, CLMs can be precisely fine-tuned for specialized tasks, exemplified by argument classification in our study. A notable benefit of CLMs is their optimization for minimized computational and energy demands. In the presented approach, the LLM plays an 'expert' role and is engaged when dealing with a subset of arguments classified with low certainty. Our research demonstrates the extent to which this combined model can enhance the outcomes of argument classification.


We formulated the following Research Questions:

\textbf{RQ1. What efficiency can the compact language model achieve across different argument classification benchmarks?}

\textbf{RQ2. How Large Language Model can improve the compact model efficiency in argument classification?}

\textbf{RQ3. What efficiency can be achieved by a hybrid model, which consists of the compact language model as a base model and the Large Language model as its fine-tuning model?}










\section{Related works}
\label{sec:headings}

Argument classification (AC) is a specific subtask within AM. It focuses on classifying the identified components of an argument into predefined categories, such as distinguishing between claims and premises or identifying the stance of an argumentative component (supporting or opposing). Argument classification can also involve determining the type of argument or its quality (e.g., strong vs. weak argument). This task is crucial for building the structured representation of arguments, as it helps in understanding the role of each component within the argument and how different components interact \cite{ref_article3}.

Various data science techniques based on natural language processing have shown efficacy in AC and AM. In the early works, argument structures were mapped through trees or similar hierarchical structures, facilitating computation with existing tree-parsing methods. However, such idealized structures seldom match the complexity of arguments encountered in practice. More recently, the focus shifted towards utilizing non-hierarchical frameworks. Prior to the advent of BERT and Transformer-based methodologies, the field of AM heavily relied on Support Vector Machines (SVMs) and neural networks for their pattern recognition and classification strengths to pinpoint and dissect argumentative text structures. Techniques like Recurrent Neural Networks, Convolutional Neural Networks, and Long Short-Term Memory units were crucial in incorporating contextual data into machine learning decision processes \cite{ref_article9}. The introduction of the first non-hierarchical model for AM by Niculae et al. \cite{ref_article10}, using factor graphs alongside structured SVMs and Bidirectional LSTM, marked a significant shift. SVMs, known for their binary classification prowess, fit well with AM tasks such as discerning argumentative elements (e.g., distinguishing claims from non-claims). Galassi et al. \cite{ref_article11} then applied LSTMs with residual connections to better predict argument components' interrelations. Subsequently, Morio et al. \cite{ref_article12} proposed using Task-Specific Parameterization and Proposition-Level Biaffine Attention for improved prediction of complex argument structures.

While SVMs and neural networks significantly contributed to the development of AM, they had limitations, such as the need for extensive feature engineering and difficulties in capturing long-distance dependencies in text. The emergence of BERT and other Transformer-based models revolutionized the landscape of AM. Introduced by Vaswani et al. \cite{ref_vaswani}, the transformer model, with its self-attention mechanism, addressed the computational and memory efficiency issues faced by RNNs. The utility of transformers, particularly BERT \cite{ref_devlin}, in AM has spurred extensive research, with fine-tuning these models showing significant improvements across diverse tasks. Recent research has applied Transformer-based models to AM, with Reimers et al. \cite{ref_reimers} improving argument classification and clustering using BERT and ELMo, and Chakrabarty et al. \cite{ref_chakrabarty} advancing argument component and relation identification in persuasive texts. Further, Chen \cite{ref_chen} applies BERT for mapping argument structures as directed acyclic graphs, while Ruiz-Dolz et al. \cite{ref_ruiz} evaluate the performance of various Transformer-based models across domain-specific datasets to identify the most versatile model.

This paper proposes a novel AM methodology that leverages the compact transformer-based models (BERT) with ChatGPT architecture, for automatic argument classification. Distinguishing our work from predecessors, we engage with a wider array of datasets and tasks, and discuss methods to overcome challenges and boost AM efficiency through BERT usage with ChatGPT-4 based refinement.

\section{Datasets}

One of the primary contributions of our approach is the broad range of argument datasets considered. The previous studies by Ajjour et al. \cite{ref_ajjour}, Bar-Haim et al. \cite{ref_barhaim}, Boltuzic and Snajder \cite{ref_boltuzic}, Cabrio and Villata \cite{ref_cabrio}, Carstens and Toni \cite{ref_carstens}, Park and Cardie  \cite{ref_article4}, Stab et al. \cite{ref_article7}, Visser et al. \cite{ref_visser} focus on a limited number of datasets. In each of these studies, only one dataset was examined.

In our research, we decided to conduct a comparative study using three different datasets containing arguments corpora: the US2016 corpus, the Moral Maze multi-domain (UKP) corpus, and the Args.me corpus. All these datasets have gained a reputation in argument mining research undertaken in recent years, but have not yet been studied comparatively and using the latest models such as Google T5 and BERT, as well as GPT and OpenLlama technologies.

First of the processed datasets, the US2016 is an argument annotated corpus of the electoral debate carried out in 2016 in the United States. It contains both, transcriptions of the different rounds of TV debate, and discussions from the Reddit forums as detailed in Visser et al. \cite{ref_visser}. Four annotators, trained in the use of Inference Anchoring Theory \cite{ref_budzynska}, annotated the US2016 corpus with Argumentative Discourse Units (ADUs). The ADUs used in this annotation are relations occurring between propositions, 
specifically: (1) The annotations of Inference (RA), reflecting that a proposition is meant to supply a reason for accepting another proposition; (2) The annotations of Conflict (CA), reflecting that a proposition is meant to be incompatible with another proposition or relation; (3) The annotations of Rephrases (MA), reflecting that a proposition is meant to be a reformulation of another proposition; and
(4) the annotations of No Relation (NO) mean that there is no relationship between the propositions.
Table 1 presents the class distribution of the processed US2016 corpus. This dataset has a total of approximately 12.5 thousand arguments.

\begin{table}[H]
\centering
\begin{tabular}{|c|c|c|c|c|}
\hline
\textbf{} & \textbf{US 2016} & \textbf{train}  & \textbf{test} \\ \hline
\textbf{RA}     &     2744     &    2470                            &  274 \\ \hline
\textbf{CA}     &      888     &    800                             &  88 \\ \hline
\textbf{MA}     &     705    &   635                              &  70 \\ \hline
\textbf{NO}     &      8055       &   7250                              &  805 \\ \hline
\textbf{Total}     &     12392      &                           11243      &  1149 \\ \hline

\end{tabular}
\caption{US2016 database statistics}
\label{statistc_of_datasets}
\end{table}

The second dataset used in our study was UKP \cite{ref_stab18}. It contains a corpus of arguments from online comments on eight controversial issues: abortion, cloning, death penalty, gun control, minimum wage, nuclear energy, school uniforms, marijuana legalization. The corpus includes over 25,000 instances. In Table 2 and 3 the statistics of UKP datasets are shown. The sentences were annotated by 7 people working independently, recruited through the Amazon Mechanical Turk (AMT) crowdsourcing platform. For each classification, it was necessary to achieve agreement, as measured by Cohen’s kappa ($\kappa$), at the level of 0.723; this exceeds the commonly used threshold of 0.7 for assuming the results are reliable \cite{ref_carletta}. The labels used in the classification were: (1) supporting argument; (2) opposing argument; and (3) non-argument.

\begin{table}[H]
\centering
\begin{tabular}{|c|c|c|c|c|c|}
\hline
\textbf{} & \textbf{abortion} & \textbf{cloning}  & \textbf{death penalty} &  \textbf{marijuana legalization} \\ \hline
\textbf{NON}     &      2282     &   1472                            &   1918 & 1160 \\ \hline
\textbf{PRO}     &       634         &     702                            &   424  & 535\\ \hline
\textbf{CON}     &       766        &     825                            &  1036  & 574\\    \hline

\end{tabular}
\caption{UKP database statistics part 1}
\label{statistc_of_ukp}
\end{table}

\begin{table}[H]
\centering
\begin{tabular}{|c|c|c|c|c|c|}
\hline
\textbf{} & \textbf{gun control} & \textbf{minimum wage}  & \textbf{nuclear energy} & \textbf{school uniforms} \\ \hline
\textbf{NON}     &      1846     &   1323                             &   2051 & 1734 \\ \hline
\textbf{PRO}     &       775          &     564                             &   591  & 545\\ \hline
\textbf{CON}     &       650        &     541                            &  831  & 729\\    \hline

\end{tabular}
\caption{UKP database statistics part 2}
\label{statistc_of_ukp}
\end{table}

The last dataset used in our study was Args.me corpus (version 1.0, cleaned), provided by Ajjour et al. \cite{ref_ajjour}. The distribution of this corpus is presented in Table 4. It consists of arguments crawled from four debate portals in the middle of 2019. The debate portals are Debatewise, IDebate.org, Debatepedia, and Debate.org. The arguments were extracted using heuristics that are designed for each debate portal. Idebate.org, Debatepedia, and Debatewise are the sub-datasets used in our simulations. These three sub-datasets consist of 48 798 arguments. The Debatepedia has the highest number of arguments. In IDebate.org the disproportion between PRO and CON arguments is the lowest, and the highest disproportion appears in Debatepedia. Args.me is the newest and the most extensive dataset among all datasets utilized in our study.
Table 4 presents the statistics of the Args.me dataset. The annotations used in this database include conclusions and premises, which are divided into PRO premises (arguments supporting the conclusion) and CON premises (arguments against the conclusion).







\begin{table}[H]
\centering
\begin{tabular}{|c|c|c|c|c|}
\hline
\textbf{} & \textbf{idebate.org} & \textbf{debatepedia}  & \textbf{debatewise} \\ \hline
\textbf{conclusions}     &      5011        &       10314                         &  5992 \\ \hline
\textbf{premises}     &     13248            &      21197                            &  14353 \\ \hline
\textbf{PRO premises}     &     6701           &      15791                            &   8514 \\ \hline
\textbf{CON premises}     &     6547            &     5406                            &   5839 \\ \hline

\end{tabular}
\caption{Args.me database statistics}
\label{statistc_of_datasets}
\end{table}



\section{Method}
\subsection{Fine tuning the language model}
The compact language models which were used in following experiments are DistilBERT and BERT models. The classifiers were built by adding fully connected layers at the output of these to transformer based models. 


Each Transformer based language model consists of the list of Transformers blocks \cite{ref_li}: 
\begin{equation}
T_{\Theta}(X) = {t_{\theta_L}(t_{\theta_{L-1}}...(t_{\theta_{0}}(X)))}
\label{eq:model}
\end{equation}

the $\Theta$ is a list:
\begin{equation}
\Theta = \{\theta_{0}, \theta_{1},...,\theta_{L}\}
\label{eq:theta_}
\end{equation}

is a list of parameters $\theta_{i}$ for each Transformer block. Each tensor $\theta_{i}$ is a set of Transformer block FC layers attention weights. The $T_{\Theta}$ gives an output with the same number of tokens as in input $X$ (generates embedding for each token). The classifiers in the presented experiments are defined as follows:

\begin{equation}
C_{\Theta_c}(X) = f_{\theta_{2}}(f_{\theta_{1}}(T_{\Theta}(X)))
\label{eq:model}
\end{equation}

The two fully connected layers are added at the top of the Transformer blocks. The output size of the last fully connected layer is equal to the number of classes to recognize.

\section{ChatGPT-4 based refinement}

The next stage of our research entailed the application of the ChatGPT model developed by OpenAI. The primary objective was to ascertain the feasibility of enhancing the efficacy of argument classification through the utilization of the GPT-4 model, particularly in instances where the BERT model exhibited limitations. 

The GPT-4 refinement algorithm is presented in Algorithm \ref{alg:refinement}. The algorithm involves creating a classifier based on a pre-trained language model as described in the previous chapter (line 1). Then the model created in this way is fine-tuned on the dataset (lines from 2 to 6). In the next step, the threshold is calculated based on the uncertainty of the model's response (lines from 8 to 11). The arguments with the least certain answer are then routed to the large language model (lines 13 and 14). The remaining arguments are processed by a small language model (line 16).

\begin{algorithm}
\begin{algorithmic}[1]
\REQUIRE{$T_{\Theta}$, $LLM$, $T$-train data, $V$-validation data, $\gamma$-threshold}
\STATE{$C_{\Theta_c}$ $\gets$ $f_{\theta_{2}}(f_{\theta_{1}}(T_{\Theta})$  //add fully connected layers}
\FOR{$b$ $\textbf{in}$ $T$}
\STATE{$L_C$ $\gets$ Loss($C_{\Theta_c}(X_b)$, $Y_b$) }
\STATE{$g$ $\gets$ grad($L_C$)}
\STATE{update $\Theta_c$}
\ENDFOR
\STATE{$\Lambda$ $\gets$ $\emptyset$}
\FOR{$b$ $\textbf{in}$ $T$}
\STATE{$\Lambda \gets \Lambda \cup  |C_{\Theta_c}(X_b)$-$Y_b|$}
\ENDFOR
\STATE{$\gamma$ $\gets$ $\Lambda[kth]$}
\FOR{$b$ $\textbf{in}$ $V$}
\IF{$|C_{\Theta_c}(X_b)$-$Y_b| > \gamma$}
\STATE{$Y \gets LLM(X_b)$}
\ELSE
\STATE{$Y \gets C_{\Theta_c}(X_b)$}
\ENDIF
\ENDFOR

\caption{Small language model with ChatGPT-4 refinement}
\label{alg:refinement}
\end{algorithmic}
\end{algorithm}
The subset of arguments from all three databases was evaluated with ChatGPT-4 when running Algorithm \ref{alg:refinement}. In each case, the same simple input-output prompting template was used (theses and arguments were inserted into blank spaces):
\begin{quote}
This is a thesis “...”. And this is a claim “...”. Is the claim an argument for or contra the thesis? Write a one-sentence answer.
\end{quote}

The responses provided by ChatGPT-4 were predominantly accurate, enhancing the performance quality of the BERT model. Nonetheless, ChatGPT-4 was not exempt from errors, the analysis of which may yield intriguing insights. The analysis of erroneous outputs indicates that the model, in some cases, struggles with processing logical connectives. In one such instance, when asked to elaborate on how it arrived at an inaccurate answer, ChatGPT-4 responded as follows: "I came to this conclusion because the thesis statement argues that Tunisia should not rely on tourism for economic growth, while the claim argues that tourism causes pollution, which suggests that relying on tourism may have negative consequences, supporting the argument against the thesis". Furthermore, ChatGPT-4 fails to recognize sarcasm. In one example, a negative argument was presented with a tone of irony, which confused the model and led to the incorrect classification of the argument as positive. There were also cases in which the model openly contradicted itself when questioned after providing an incorrect answer. 

Another type of error was associated with a failure to understand arguments that place the issue in a context different from the most commonly encountered one. For example, in a debate about cloning where the assertion was 'Cloning should be allowed,' ChatGPT-4 did not grasp an argument concerning the positive impact of cloning on “creating plants that offer better nutritional value”. Consequently, given that – as it articulated – “debates on cloning typically focus on the ethical, legal, or social implications of cloning humans or animals”, ChatGPT-4 claimed that this statement did not contain any argument. 

A frequently recurring problem is also the incorrect classification of arguments expressed in sentences containing multiple negations. For instance, the sentence 'The fact that some states or countries which do not use the death penalty have lower murder rates than jurisdictions which do is not evidence of the failure of deterrence' was classified as an argument against the death penalty, despite the explicit clarification that it is 'not evidence'. Moreover, there are instances where ChatGPT-4 contradicts itself when justifying an incorrect response, claiming, for example, that 'the statement implicitly argues against the proposition that "Gun access should not be limited" (...) Therefore, it is an argument against the proposition that gun access should be limited.

Finally, we observed that some datasets, in particular UKP, contain a noticeable group (approximately 10\%) of incorrect or problematic classifications made by human evaluators, while ChatGPT's evaluation of these cases seems to be more consistent with reality.

\section{Results and discussion}

\begin{table}
\centering
\begin{tabular}{|c|c|c|c|c|c|}
\hline
\textbf{} & \textbf{idebate.org} & \textbf{debatepedia}  & \textbf{debatewise} & \textbf{top1} & \textbf{F1}\\ \hline
\textbf{DistilBERT}     &    60.63           &      72.64     &   60.02  &  65.1 &  68.9\\ \hline
\textbf{BERT}     &     73.76             &      83.1      &     69.2  &  76.22 &  71.5 \\ \hline
\textbf{BERT + ChatGPT-4}     &        84.82         &                 91.41                &    76.12 & 85.87 & 
 87.2 \\ \hline

\end{tabular}
\caption{Top1 and F1 Accuracy for Args.me - PRO and CON arguments recognition without conclusion.}
\label{argsme}
\end{table}

\begin{table}
\centering
\begin{tabular}{|c|c|c|c|c|c|c|}
\hline
\textbf{} & \textbf{idebate.org} & \textbf{debatepedia}  & \textbf{debatewise} & \textbf{top1} & \textbf{F1} \\ \hline
\textbf{DistilBERT}     &      60.48       &    72.83                     &  60.02 &   65.1 & 69.3\\ \hline
\textbf{BERT}     &      83.1           &                  85.48              &  84.3 &  84.22 &  85.3 \\ \hline
\textbf{BERT + ChatGPT-4}     &        88.91        &                  92.92              &    88.51 & 89.6 & 91.3 \\ \hline
Akiki et al. \textbf{\cite{akiki_2020}}     & - & - & - & 75.5 & - \\ \hline

\end{tabular}
\caption{Top1 and F1 Accuracy for Argsme - PRO and CON arguments recognition with conclusion}
\label{argsme_context}
\end{table}

\begin{table}
\centering
\begin{tabular}{|c|c|c|}
\hline
\textbf{} & \textbf{US2016} \\ \hline
\textbf{LSTM \cite{ref_li}}     &   26.0       \\ \hline
\textbf{DistilBERT \cite{ref_li}}     &   55.0        \\ \hline
\textbf{BERT \cite{ref_li}}     &     62.0   \\ \hline
\textbf{BERT + ChatGPT-4}     &     72.5  \\ \hline

\end{tabular}
\caption{F1 metric for US2016 (four labels)}
\label{us2016}
\end{table}

\begin{table}
\centering
\begin{tabular}{|c|c|c|c|c|c|}
\hline
\textbf{} & \textbf{abortion} & \textbf{cloning}  & \textbf{death penalty} & \textbf{marijuana legalization}\\ \hline
\textbf{DistilBERT}     &     60.0  &    51.33     &  57.39 &  54.38 \\ \hline
\textbf{BERT}     &       73.51        &   70.66       &  65.68 & 61.40 \\ \hline
\hline
\textbf{BERT + ChatGPT-4}     &       77.18        &               74.19       &   74.19  &  64.47 \\ \hline

\end{tabular}
\caption{Top1 Accuracy for UKP dataset, part 2 (three labels).}
\label{ukp_1}
\end{table}

\begin{table}
\centering
\begin{tabular}{|c|c|c|c|c|c|}
\hline
\textbf{} & \textbf{gun control} & \textbf{minimum wage}  & \textbf{nuclear energy} & \textbf{school uniforms} \\ \hline
\textbf{DistilBERT}     &   60.36  &    59.01                           &  63.79  &  52.98 \\ \hline
\textbf{BERT}     &      70.12         &   68.03                             &  77.01  & 76.15 \\ \hline
\textbf{BERT + ChatGPT-4}     &      75.02     &   72.79    &   80.86   &  79.95 \\ \hline

\end{tabular}
\caption{Top1 Accuracy for UKP dataset, part 2 (three labels).}
\label{ukp_2}
\end{table}

\begin{table*}[!ht]
\centering
\begin{tabular}{|c|c|c|}
\hline
\textbf{model} & \textbf{UKP} \\ \hline
\textbf{LSTM \cite{ref_stab18}}     &    42.85      \\ \hline
\textbf{DistilBERT}     &     54.2     \\ \hline
\textbf{BERT}     &    57.7   \\ \hline
\textbf{BERT + ChatGPT-4}     &   68.5   \\ \hline

\end{tabular}
\caption{Mean F1 score for UKP (three labels)}
\label{ukp+f1}
\end{table*}

Table \ref{argsme} and Table \ref{argsme_context} show the results of all conducted evaluations for the Args.me dataset. Trainings for the DistilBERT and BERT models were carried out on three subsets: debatepedia, idebate and debatewise. One experiment involved providing only the argument text in training and evaluation as input. However, in the second experiment, an argument and a thesis were given as input. As can be observed, better results were achieved in the second case. For all three subsets the improvement in both top1 and F1 metrics can be observed (the highest improvements were achieved for idebate.org and debatewise). The best results were obtained for the debatepedia dataset (91.41 and 92.92 top1. \textbf{RQ3}). Among all models (BERT, DistilBERT and \cite{akiki_2020}) BERT+ChatGPT-4 is the best and outperforms other models (by 16\% in F1 for the first case and by 6\% for the second one, \textbf{RQ2}).  
The $\gamma$ parameter for Args.me dataset was set to value that about 20\% of arguments with the highest uncertainty were delegated to ChatGPT-4.

Table \ref{us2016} shows the results for the US21016 set. The model for this set classifies the input data into four possible classes. The results for the BERT model (the F1 metric is 62.0) are definitely better than for the DistilBERT (55.0) and LSTM (26.0) models (\textbf{RQ1}). The reported results are the same as presented in \cite{ref_li}. In the case of the BERT+ChatGPT-4 mode, the $\gamma$ parameter from Algorithm \ref{alg:refinement} was set so that 25\% of the data with the lowest confidence was delegated from the BERT model to the ChatGPT-4 model. The BERT+ChatGPT-4 again achieves the best results (72.5\%, \textbf{RQ3}). The F1 metric is improved by about 10\% (\textbf{RQ2}).

Table \ref{ukp_1} and Table \ref{ukp_2} present the results obtained for the UKP dataset. The top1 metric for each thematic subset (8 subsets) is presented. As can be seen, the results are the best for the largest subsets (especially abortion and nuclear energy). This is due to the fact that the model has a larger number of training courses, which allows it to better perform the fine-tuning process. As expected, BERT achieves better results than DistilBERT in all groups. However, the BERT+ChatGPT-4 hybrid model outperforms both of them. It should be mentioned that during the analysis of the model's response, the cases were noticed in which the labels of some arguments were incorrect or controversial. This may also result in the lowest accuracy of UKP across all datasets. The second factor is the amount of data in this dataset.

Finally, Table \ref{ukp+f1} shows the aggregated results of the UKP set for the F1 metric. The correlation of results between models is maintained. The BERT+ChatGPT-4 outperforms all presented models (by at least 10\%, \textbf{RQ2}) and achieves 68.5\%(\textbf{RQ3}). The BERT and DistilBERT outperform again LSTM-based model \cite{ref_stab18} as in the case of US2016, \textbf{RQ1}.

One of the main goals of the work was to show the possibility of increasing the results of a compact language model by adding a large language model as an expert in selected cases (in parallel minimizing the usage of LLM). Thanks to this, you can scale the inference time (model response) and reduce the use of a large language model. 

All our experiments were executed on a workstation equipped with Nvidia Tesla V100-SXM2-32GB GPUs.

\section{Conclusions and future works}

The presented results show that compact language models such as BERT and Google T5 do not achieve fully satisfactory results in argument classification. Their advantage, however, is their speed of fine-tuning. Limited access to large models and their computational complexity makes it difficult to evaluate the collections quickly. Our study shows that the results of small models can be easily and quickly improved by using large models for a selected subset of arguments.
Our research also shows that argument-mining datasets have minor shortcomings in the form of incorrect labeling.

Future work will focus on large language models. In addition to the ChatGPT-4 model, open-source models will be explored, including LLAMA-2. In order to conduct further research in GPT-based refinement, it will be necessary to utilize more advanced prompting techniques, such as the Tree of Thoughts, which recent reports suggest can significantly enhance ChatGPT's reasoning capabilities \cite{ref_yao}.

\end{document}